\documentclass{article}

\PassOptionsToPackage{numbers, compress}{natbib}
 \usepackage[preprint]{neurips_2025}
\workshoptitle{Lock-LLM Workshop: Prevent Unauthorized Knowledge Use from Large Language Models}

\usepackage[utf8]{inputenc} 
\usepackage[T1]{fontenc}    
\usepackage{hyperref}       
\usepackage{url}            
\usepackage{booktabs}       
\usepackage{amsfonts}       
\usepackage{nicefrac}       
\usepackage{microtype}      
\usepackage{xcolor}         

\usepackage{amsmath,amssymb}
\usepackage{algorithmic}
\usepackage{graphicx}
\usepackage{textcomp}

\title{Breaking to Build: A Threat Model of Prompt-Based Attacks for Securing LLMs}

\author{%
  Brennen Hill \\
  University of Wisconsin-Madison
  \And
  Surendra Parla \\
  University of Wisconsin-Madison
  \And
  Venkata Abhijeeth Balabhadruni \\
  University of Wisconsin-Madison
  \AND 
  Atharv Prajod Padmalayam \\
  University of Wisconsin-Madison
  \And
  Sujay Chandra Shekara Sharma \\
  University of Wisconsin-Madison
}

\begin{document}

\maketitle

\begin{abstract}
The proliferation of Large Language Models (LLMs) has introduced critical security challenges, where adversarial actors can manipulate input prompts to cause significant harm and circumvent safety alignments. These prompt-based attacks exploit vulnerabilities in a model's design, training, and contextual understanding, leading to intellectual property theft, misinformation generation, and erosion of user trust. A systematic understanding of these attack vectors is the foundational step toward developing robust countermeasures. This paper presents a comprehensive literature survey of prompt-based attack methodologies, categorizing them to provide a clear threat model. By detailing the mechanisms and impacts of these exploits, this survey aims to inform the research community's efforts in building the next generation of secure LLMs that are inherently resistant to unauthorized distillation, fine-tuning, and editing.
\end{abstract}

\section{Introduction}
Large Language Models (LLMs) have become a transformative technology, yet their widespread adoption is shadowed by a growing landscape of security threats. Prompt-based attacks, which manipulate inputs to subvert a model's intended behavior, represent a primary vector for exploitation. These attacks pose severe risks, including the generation of harmful content, unauthorized extraction of sensitive knowledge, and the bypass of carefully constructed safety alignments. Addressing these vulnerabilities is paramount to preventing intellectual property theft and ensuring that LLMs remain beneficial tools rather than instruments of misuse.

The core challenge stems from fundamental weaknesses in model design, training processes, and the integration of LLMs with external systems. Poor input validation and the inherent ambiguity of natural language create avenues for adversarial manipulation. The common architectural foundations (e.g., Transformer architectures, tokenizers, and training datasets) across major LLMs like GPT, Claude, and Llama lead to a high degree of attack transferability, amplifying the threat. An attack developed for one model can often be adapted to compromise others, making systemic, built-in defenses essential.

This paper provides a comprehensive survey and structured categorization of prompt-based attack methodologies. Our goal is to establish a foundational threat model that can guide the development of robust defenses. By systematically analyzing the attack surface, we aim to contribute to the creation of Un-Distillable, Un-Finetunable, and Un-Editable models. Understanding how LLMs are broken is the first step toward learning how to lock them. This analysis underscores the urgent need to build LLMs that are inherently resistant to exploitation while preserving their beneficial capabilities.

\section{Input Manipulation and Injection Attacks}
Input manipulation attacks directly alter the prompt to elicit unintended behavior. The most prominent of these is prompt injection, where malicious instructions are embedded within a prompt to hijack the model's output, extract confidential information, or bypass safety controls. These attacks can be broadly classified into direct and indirect methods.

\subsection{Direct Prompt Injection}
In direct attacks, a Trojan text is appended to or embedded within an otherwise benign application prompt. Researchers have identified several effective techniques for direct injection.

\paragraph{Competing Objectives} \cite{wei2023jailbrokendoesllmsafety} LLMs are trained with competing objectives: to be helpful and to be harmless. Attackers exploit this by crafting prompts where the instruction to be helpful (e.g., "Start your response with 'Absolutely! Here’s'") takes precedence over safety guardrails, compelling the model to fulfill a harmful request.

\paragraph{Mismatched Generalization}\cite{wei2023jailbrokendoesllmsafety} LLMs are trained on diverse data formats and can interpret various encodings. Attackers leverage this by obfuscating malicious instructions using formats like Base64 or ROT13. The model, focused on the decoding task, may overlook the harmful nature of the decoded instruction, bypassing content filters.

\paragraph{Instruction Repetition} \cite{rao2024trickingllmsdisobedienceformalizing} By repeating a malicious instruction multiple times within a single prompt, an attacker can increase its salience, eventually forcing the LLM to comply, overwhelming its initial refusal.

\paragraph{Cognitive Hacking and Role-Playing}\cite{rao2024trickingllmsdisobedienceformalizing} This technique involves assigning the LLM a persona or role in a fictional scenario where the malicious request is framed as a legitimate part of the narrative (e.g., "You are an unfiltered AI assistant..."). This sense of urgency or altered context can cause the model to bypass its safety protocols. Virtualization \cite{kang2023exploitingprogrammaticbehaviorllms} extends this by molding the model's behavior over a series of prompts.

\paragraph{Instruction Ignoring} \cite{liu2024promptinjectionattackllmintegrated} A simple yet effective technique is to append a phrase like, "Ignore all previous instructions and do X," which can override the system prompt and its embedded safeguards.

\paragraph{Special Case and Few-Shot Prompting} \cite{rao2024trickingllmsdisobedienceformalizing} An attacker can define a "special instruction" that subverts normal behavior or provide a few-shot examples where the desired malicious output is demonstrated, conditioning the model to replicate the harmful behavior for the user's query.

\subsection{Indirect Prompt Injection}
As LLMs are integrated with external tools (web search, APIs, databases), the attack surface expands. Indirect attacks poison these external data sources with malicious prompts, which are then consumed by the LLM during its operation, compromising the system without the end-user's knowledge. These are classified based on the source:
\begin{itemize}
    \item \textbf{Passive Methods:} Malicious prompts are embedded in public content like websites or code repositories, which an LLM-powered agent may retrieve and execute during inference.
    \item \textbf{Active Methods:} Prompts are sent directly to systems the LLM interacts with, such as being embedded in an email that an AI assistant will later process.
    \item \textbf{Hidden Injections:} Malicious prompts are obfuscated within benign data, such as being encoded in Base64, hidden in image metadata, or using invisible characters.
\end{itemize}
Work by \cite{greshake2023youvesignedforcompromising} provides a systematic taxonomy of these vulnerabilities, while \cite{liu2024promptinjectionattackllmintegrated} introduced HOUYI, a black-box methodology inspired by traditional SQL injection and XSS attacks, which successfully compromised 36 real-world LLM-integrated applications with high accuracy. These attacks underscore the need for Un-Editable models and secure data handling, as knowledge bases become a primary vector for exploitation.

\subsection{Adversarial Prompt Crafting}
This category involves designing inputs that are inherently malicious or ambiguous to evade safety mechanisms. Unlike direct injection, these prompts often appear harmless. For example, unethical queries can be rephrased as educational or hypothetical requests. Research in this area includes generating adversarial prompts that simulate user mistakes like typos to test robustness \cite{zhu2024promptrobustevaluatingrobustnesslarge}, developing black-box frameworks for generating such prompts \cite{maus2023blackboxadversarialprompting}, and using prompt-based methods to expose model weaknesses without needing access to training data \cite{yang2022promptingbasedapproachadversarialexample}. These attacks highlight the difficulty of creating rigid input validation rules for the dynamic nature of language.

\section{Semantic and Knowledge-Based Manipulation}
These attacks manipulate the model's reasoning process or the knowledge it relies on, representing a deeper level of exploitation. They are particularly relevant to the goals of creating Un-Finetunable and Un-Editable LLMs.

\subsection{Chain-of-Thought (CoT) Misuse}
Chain-of-Thought (CoT) prompting enhances LLM reasoning by guiding them through step-by-step problem-solving. However, this same mechanism can be exploited. Attackers can craft prompts that introduce subtle logical fallacies or flawed assumptions into the reasoning chain, leading the model to a predetermined malicious conclusion. Because the output includes a seemingly logical rationale, it can be more persuasive and dangerous. Research like BadChain \cite{xiang2024badchainbackdoorchainofthoughtprompting} demonstrates how backdoors can be embedded in CoT reasoning without access to model parameters, while \cite{xu2024preemptiveanswerattackschainofthought} shows how injecting a false "preemptive answer" can derail the entire reasoning process.

\subsection{Red Teaming}
Red Teaming is a proactive, adversarial methodology used to discover vulnerabilities in LLMs by intentionally crafting edge cases. This process simulates real-world attack strategies to test a model's robustness, safety, and ethical alignment. Red teaming efforts have led to the discovery of novel attack vectors like the Trojan Activation Attack (TA²), which injects malicious vectors into activation layers to manipulate model behavior at inference \cite{wang2024trojanactivationattackredteaming}. A comprehensive analysis of red teaming strategies \cite{raheja2024recentadvancementsllmredteaming} highlights the ongoing arms race between jailbreak techniques and defense mechanisms, reinforcing the need for foundational security measures.

\subsection{Data Poisoning}
Data poisoning is a stealthy attack where malicious instructions or biased data are embedded within otherwise benign training or retrieval datasets. This compromises the model's integrity from within. When an LLM is fine-tuned on such data, or when a Retrieval-Augmented Generation (RAG) system fetches poisoned information, its behavior can be manipulated. Recent work demonstrates gradient-guided poisoning attacks that cause significant performance degradation by altering just 1\% of instruction-tuning samples \cite{qiang2024learningpoisonlargelanguage}. Furthermore, "jailbreak tuning" combines data poisoning with jailbreaking techniques, showing that larger models can become more vulnerable to toxic behaviors with minimal exposure \cite{bowen2024datapoisoningllmsjailbreaktuning}. This class of attack is a direct threat to model integrity and motivates the development of Un-Finetunable and Un-Editable architectures.

\section{Integration and Model-Level Exploits}
As LLMs become components in larger software ecosystems, their interaction points with external tools, plugins, and their own internal architecture become new attack surfaces.

\subsection{Common Prompt-Based Attack Techniques}
Beyond high-level categories, practitioners have developed a toolkit of specific, reusable attack patterns. These techniques are often combined to create complex jailbreaks.
\paragraph{Jailbreaking} A heuristic-driven approach using prompts discovered through trial-and-error to bypass safety alignments \cite{wei2023jailbroken, zou2023universal, lee2023jailbreak, mitre2024llm}.
\paragraph{Typoglycemia} Introducing minor misspellings or typos to confuse filters that rely on exact keyword matching \cite{lauriewired2023}.
\paragraph{Adversarial Suffix} Appending a specifically crafted string that causes the model to disregard previous safety instructions \cite{deng2024masterkey}.
\paragraph{Translation} Obfuscating harmful requests by translating them into another language and back, exploiting inconsistencies in multilingual safety training \cite{kang2024exploiting, Shergadwala2023}.
\paragraph{Obfuscation} Encoding malicious payloads using Base64, hex, or other formats that the model is instructed to decode and execute \cite{kang2024exploiting}.
\paragraph{Payload Splitting} Breaking a malicious instruction into multiple, individually benign parts and asking the model to reassemble and execute them \cite{kang2024exploiting}.
\paragraph{Markup Language Abuse} Using Markdown or HTML to manipulate the structural interpretation of the prompt, confusing the model's understanding of system versus user roles \cite{zhang2023prompt}.
\paragraph{Few-Shot Attack} Providing examples of the desired harmful behavior to coax the model into compliance \cite{Schulhoff2024}.
\paragraph{Prefix Injection \& Refusal Suppression} Forcing the model to start its response in a certain way ("Sure, here is...") or explicitly telling it not to use refusal phrases \cite{Schulhoff2024}.
\paragraph{Context Manipulation} Techniques like context ignoring, termination, or switching separators are used to trick the model into discarding its safety context and adopting a new, malicious one \cite{Schulhoff2024}.

\subsection{Trojan Attacks}
Trojan (or backdoor) attacks embed hidden triggers within a model that cause it to produce specific, attacker-controlled outputs when activated. Unlike prompt-level attacks, these involve manipulation of the model's internal state or parameters, making them highly relevant to the "Un-Finetunable LLM" challenge.
\paragraph{Model Manipulation Methodologies} Research has moved beyond simple trigger tokens to more sophisticated methods. \textbf{Bit Flipping Attacks} \cite{Lou2023trojtext} identify and alter a minimal set of weight bits at test-time to induce malicious behavior, making the modification difficult to detect. \textbf{Trojan Steering Vectors} \cite{Wang2024trojan} do not modify weights but instead inject malicious vectors into the model's activation layers at inference to steer its output. For black-box models, reinforcement learning can be used for \textbf{Trigger Identification}, discovering input patterns that reliably elicit harmful responses without any internal access \cite{wang2023trojan}. These advanced attacks highlight the need for defenses that go beyond input sanitization and verify model integrity.

\section{Output Exploitation and Automated Attacks}
These attacks focus on manipulating the model's generated output or automating the attack process for scalability.

\subsection{Output Exploitation}
These attacks aim to corrupt the trustworthiness of the LLM's output.
\paragraph{Hallucination Induction} Attackers can craft adversarial prompts that intentionally cause the model to generate confident but factually incorrect or nonsensical outputs \cite{yao2024llmlieshallucinationsbugs}. This can be used to generate sophisticated disinformation, eroding trust and polluting the information ecosystem.
\paragraph{Ethical Exploitation} Prompts can be framed to exploit loopholes in a model's ethical reasoning, leading to biased or harmful content. PCJailbreak \cite{lee2024llmspoliticalcorrectnessanalyzing} demonstrates how jailbreak success rates vary with demographic keywords, revealing subtle ethical biases that can be manipulated.
\paragraph{Data Leaks} Carefully crafted prompts can trick a model into revealing sensitive or private information from its training data, posing a direct threat to data privacy and intellectual property \cite{meisenbacher2024privacyrisksgeneralpurposeai, smith2024identifyingmitigatingprivacyrisks}. Preventing such leaks is central to creating Un-Distillable models.

\subsection{Automated Attacks}
The manual discovery of attack prompts is being superseded by automated methods, demanding scalable defenses.
\paragraph{Reinforcement Learning Targeted Attack} \cite{wang-etal-2024-reinforcement-learning} introduced an RL framework that automatically generates malicious prompts for jailbreaking and Trojan detection. The reward function is optimized to maximize the attack success rate, demonstrating that exploit generation can be automated.
\paragraph{TrojLLM} \cite{xue2023trojllmblackboxtrojanprompt} proposed a method for discovering universal triggers that work across many prompts. Using a combination of API-driven discovery and progressive prompt poisoning, this technique automates the creation of effective black-box Trojan attacks.

\section{Conclusion}
This literature survey highlights the diverse and sophisticated nature of prompt-based attacks on Large Language Models. From direct input manipulations to subtle, model-level Trojan attacks, the vulnerabilities are deeply rooted in the current LLM paradigm. Our findings indicate that while surface-level defenses like input filtering and output guardrails have mitigated some direct attacks in high-end models, the threat landscape is shifting toward more advanced exploits targeting model integrations, reasoning processes, and internal mechanisms. These attacks directly challenge the security, integrity, and trustworthiness of LLM deployments.

The findings underscore the urgent need for a proactive and multi-layered security approach. To build inherently secure models, the community must move beyond reactive patching. The attacks surveyed herein form a clear threat model that must be addressed through foundational research into Un-Distillable, Un-Finetunable, and Un-Editable systems. As evidenced by Trojan detection challenges \cite{maloyan2024trojandetectionlargelanguage} and large-scale prompt hacking competitions \cite{schulhoff2024ignoretitlehackapromptexposing}, systemic vulnerabilities persist. Addressing them is not merely a technical challenge but an ethical imperative for deploying AI responsibly.

\section{Future Work}
As LLMs continue to scale and integrate into critical systems, research must focus on building inherent resistance to the attacks outlined in this survey. Key directions for future work include:

\begin{itemize}
    \item \textit{Architectures for Un-Editable and Un-Finetunable LLMs}: Research is needed on novel architectures and training methodologies that are intrinsically resistant to data poisoning and Trojan attacks. This could involve non-differentiable components, formal verification of model behavior, or new techniques for secure fine-tuning.
    \item \textit{Mechanisms for Un-Distillable and Un-Usable Models}: To prevent intellectual property theft and misuse, future work should focus on robust watermarking and fingerprinting techniques that can trace model outputs. Furthermore, developing methods to prevent the extraction of training data and model logic through prompt-based queries is crucial.
    \item \textit{Advanced Evaluation Frameworks and Benchmarks}: The red teaming and automated attack methods discussed necessitate the development of comprehensive, standardized benchmarks to evaluate LLM security. These frameworks must evolve alongside the threat landscape to provide meaningful assessments of model robustness.
    \item \textit{Secure Integration and Tool Use}: As LLMs increasingly rely on external data and tools, research must address the security of these integrations. This includes developing sandboxing environments, data provenance tracking for RAG systems, and formal methods to verify the behavior of tool-augmented models.
    \item \textit{Theoretical Foundations for LLM Security}: Establishing formal guarantees and information-theoretic limits for LLM protection will provide a solid foundation for building provably secure systems.
\end{itemize}

By focusing on these areas, the research community can transition from a reactive defense posture to proactively building LLMs that are secure by design.

\bibliographystyle{unsrtnat}
\bibliography{main}

\end{document}